\documentclass[runningheads]{llncs}
\usepackage[T1]{fontenc}
\usepackage{amsmath,amssymb,amsfonts}
\usepackage[linesnumbered,ruled,vlined]{algorithm2e}

\usepackage{graphicx}
\begin{document}
\title{Deep Reinforcement Learning Based Vehicle Selection for Asynchronous Federated Learning Enabled Vehicular Edge Computing}
\titlerunning{DRL Based Vehicle Selection for AFL Enabled Vehicular Edge Computing}
%
\author{Qiong Wu\inst{1,2} \and
Siyuan Wang\inst{1,2} \and
Pingyi Fan\inst{3} \and Qiang Fan\inst{4}}
\authorrunning{Q. Wu et al.}
%
\institute{School of Internet of Things Engineering, Jiangnan University, Wuxi 214122, China
\email{qiongwu@jiangnan.edu.cn, siyuanwang@stu.jiangnan.edu.cn}\\
\and
State Key Laboratory of Integrated Services Networks (Xidian University),  Xi'an 710071, China \\
\and
Department of Electronic Engineering, Beijing National Research Center for Information Science and Technology, Tsinghua University, Beijing 100084, China\\
\email{fpy@tsinghua.edu.cn}\\
\and
Qualcomm, San Jose CA 95110 USA
\email{qf9898@gmail.com}}
\maketitle              
\begin{abstract}
In the traditional vehicular network, computing tasks generated by the vehicles are usually uploaded to the cloud for processing. However, since task offloading toward the cloud will cause a large delay, vehicular edge computing (VEC) is introduced to avoid such a problem and improve the whole system performance, where a roadside unit (RSU) with certain computing capability is used to process the data of vehicles as an edge entity. Owing to the privacy and security issues, vehicles are reluctant to upload local data directly to the RSU, and thus federated learning (FL) becomes a promising technology for some machine learning tasks in VEC, where vehicles only need to upload the local model hyperparameters instead of transferring their local data to the nearby RSU.
Furthermore, as vehicles have different local training time due to various sizes of local data and their different computing capabilities, asynchronous federated learning (AFL) is employed to facilitate the RSU to update the global model immediately after receiving a local model to reduce the aggregation delay.
However, in AFL of VEC, different vehicles may have different impact on the global model updating because of their various local training delay, transmission delay and local data sizes.
Also, if there are bad nodes among the vehicles, (that is, the amount of data and local computing resources are small, and the local model is polluted by random noise), it will affect the global aggregation quality at the RSU.\renewcommand{\thefootnote}{}
\footnotetext{Supported in part by the National Natural Science Foundation of China (No. 61701197), in part by the open research fund of State Key Laboratory of Integrated Services Networks (No. ISN23-11), in part by the National Key Research and Development Program of China (No. 2021YFA1000500(4)), in part by the 111 Project (No. B23008). (Qiong Wu and Siyuan Wang contributed equally to this work.)}
To solve the above problem, we shall propose a deep reinforcement learning (DRL) based vehicle selection scheme to improve the accuracy of the global model in AFL of vehicular network. In the scheme, we present the model including the state, action and reward in the DRL based to the specific problem. Simulation results demonstrate our scheme can effectively remove the bad nodes and improve the aggregation accuracy of the global model.

\end{abstract}
\section{Introduction}

The emerging Internet of Vehicles (IoV) becomes a promising technology to make our life more convenient \cite{i1,other4,other5,other11}.
At the same time, intelligent services also become a critical part in various vehicles\cite{other7}. Therefore, vehicles driving on the road will generate some computing tasks according to the high quality service requirements of users \cite{i2,other2}.
However, in the traditional cloud computing, the cloud is far from the moving vehicles, incurring a high task delay when tasks are offloaded to the cloud, which is not suitable for high-speed vehicles.
Thus, vehicular edge computing (VEC) \cite{i3} is introduced to enable vehicles to offload the computing tasks to a roadside unit (RSU) with a certain computing capability to reduce the task processing delay. However, it requires the vehicle to upload local data to the RSU for processing which is a challenging issue because people are reluctant to open their local data due to privacy issues \cite{i4,add1}.
So the federated learning (FL) is designed to handle this issue \cite{i5,i6}.
Specifically, FL performs iterative global aggregations at the RSU. In one round, the vehicle first downloads the current global model update from the RSU, and then uses their local data for local training. The trained local model will be uploaded to the RSU. When the RSU receives all the trained local models from vehicles, it will perform the global aggregation and broadcast the updated global model to the vehicles. Then the second round is repeated until the specified number of rounds is reached. Since local data cannot be accessed at RSU, the data privacy can be ensured physically.

However, in conventional FL\cite{add3}, the RSU needs to wait for all vehicles to offload the local model before updating the global model\cite{other6}. If a vehicle has high local training delay and transmission delay, some vehicles may drive out of the coverage of the RSU and thus cannot participate in the global aggregation.
Thus, asynchronous federated training (AFL) is introduced \cite{i7,i8,other1}. Specifically, the vehicle uploads the local model after finishing one round of its local training, with which the RSU updates the global model once it receives the local model. This enables a faster update of the global model at the RSU without waiting for other vehicles.

The vehicle mobility will cause the time-varying channel conditions and transmission rates\cite{other8,other9}, and thus vehicles have different transmission delays \cite{i9,other10,other3,add2}.
At the same time, different vehicles have different time-varying computing resources and different amounts of local data, which will cause different local training delays. In AFL, vehicles upload their local models asynchronously, it is possible that the RSU has already updated the global model according to its received local models but some vehicles had not uploaded its local model yet. As a result, the local model of these vehicles are in staleness. Staleness is related to local training delay and transmission delay. Therefore, it is important to consider the impact of the above factors on the accuracy of the global model at the RSU.

In AFL, some bad nodes may exist in the network, that is, the vehicle has small available computing resources, small amount of local data, or the local model is polluted by random noise. The bad nodes can significantly affect the accuracy of the global model at the RSU \cite{i10}.
Therefore, it is necessary to select the vehicles to participate in the global aggregation. Deep reinforcement learning (DRL) may provide a way to select the proper vehicles to solve this problem\cite{other12}. Specifically, it takes action based on the current state of vehicles, and then gets the corresponding reward. After that, the next state is reached, and the above steps are repeated. Finally, the neural network can provide an optimal policy for the vehicle selection of the system.

In this paper, we have proposed an AFL weight optimized scheme which selects vehicles based on Deep Deterministic Policy Gradient (DDPG) while considering the mobility of the vehicles, time-varying channel conditions, time-varying available computing resources of the vehicles, different amount of local data of vehicles, and existence of bad nodes\footnote[1]{The source code has been released at: https://github.com/qiongwu86/AFLDDPG}. The main contributions of this paper are shown as follows:

\begin{itemize}
\item[1)]By considering the bad nodes with less local data, less available computing resources and a local model polluted by random noise, we have employed DDPG to select vehicles to participate in the AFL, so as to avoid the impact of bad nodes on the global model aggregation.
\item[2)]We consider the impact of vehicle mobility, time-varying channel conditions, the time-varying available computing resources of the vehicles, and different amount of local data of vehicles to perform a weight optimization  and select vehicles that participate in the global aggregation, thereby improving the accuracy of the global model.
\item[3)]Extensive simulation results demonstrate that our scheme can effectively remove the bad nodes and improve the accuracy of the global model.
\end{itemize}

\section{Related Works}
In the literature, there are many research works on FL in vehicular networks.

In \cite{r1}, Zhou \emph{et al.} proposed a two-layer federated learning framework based on the 6G supported vehicular networks to improve the learning accuracy.
In \cite{r2}, Zhang \emph{et al.} proposed a method using federated transfer learning to detect the drowsiness of drivers to preserve drivers' privacy while increasing the accuracy of their scheme.
In \cite{r3}, Xiao \emph{et al.} proposed a greedy strategy to select vehicles according to the position and velocity to minimize the cost of FL.
In \cite{r4}, Saputra \emph{et al.} proposed a vehicle selection method based on their locations and history information, and then developed a multi-principal one-agent contract-based policy to maximize the profits of the service provider and vehicles while improving the accuracy of their scheme.
In \cite{r5}, Yan \emph{et al.} proposed a power allocation scheme based on FL to maximize energy efficiency while getting better accuracy of power allocation strategy.
In \cite{r6}, Ye \emph{et al.} proposed an incentive mechanism by using multidimensional contract theory and prospect theory to optimize the incentive for vehicles when preforming tasks. 
In \cite{r7}, Kong \emph{et al.} proposed a federated learning-based license plate recognition framework to get a high accuracy and low cost for detecting and recognizing license plate.
In \cite{r8}, Saputra \emph{et al.} proposed an economic-efficiency framework using FL for an electric vehicular network to maximize the profits for charging stations. 
In \cite{r9}, Ye \emph{et al.} proposed a selective model aggregation approach to get a higher accuracy of global model.
In \cite{r10}, Zhao \emph{et al.} proposed a scheme combined FL with local differential privacy to get a high accuracy when the privacy budget is small.
In \cite{r11}, Li \emph{et al.} proposed an identity-based privacy preserving scheme to protect the privacy of vehicular message. It can reduce the training loss while increasing the accuracy.
In \cite{r12}, Taïk \emph{et al.} proposed a scheme including FL and corresponding learning and scheduling process to efficiently use vehicular-to-vehicular resources to bypass the communication bottleneck. This scheme can effectively improve the learning accuracy.
In \cite{r13}, Hui \emph{et al.} proposed a digital twins enabled on-demand matching scheme for multi tasks FL to address the two-way selection problem between task requesters and RSUs. 
In \cite{r14}, Liu \emph{et al.} proposed an efficient-communication approach, which consists of the customized local training strategy, partial client participation rule and a flexible aggregation policy to improve the test accuracy and average communication optimization rate.
In \cite{r15}, Lv \emph{et al.} proposed a blockchain-based FL scheme to detect misbehavior and finally get higher accuracy and efficiency.
In \cite{r16}, Khan \emph{et al.} proposed a DRL-based FL to minimize the cost considering packet error rate and global loss.
In \cite{r17}, Samarakoon \emph{et al.} proposed a scheme considering joint power and resource allocation for ultra-reliable low-latency communication in vehicular networks to keep a high accuracy while reducing the average power consumption and the amount of exchanged data.
In \cite{r18}, Hammoud \emph{et al.} proposed a horizontal-based FL, empowered by fog federations, devised for the mobile environment to improve the accuracy and service quality of IoV intelligent applications.

However, these works have not considered the situation that vehicles may usually drive out of the coverage of the RSU before they upload their local models, which deteriorates the accuracy of the global model.

A few works have studied the AFL in vehicular networks. In \cite{r19}, Tian \emph{et al.} proposed an asynchronous federated deep Q-learning network to solve the task offloading problem in vehicular network, then designed a queue-aware algorithm to allocate computing resources.
In \cite{r20}, Pan \emph{et al.} proposed a scheme using AFL and deep Q-learning algorithm to get the maximized throughput while considering the long-term ultrareliable and low-latency communication constraints. However, these works have not considered the mobility of vehicles, the amount of data and computing capability to select vehicle in the design of the AFL in vehicular networks and the impact of bad nodes.
This motivates us to do this work by considering the key factors affecting the AFL applications in vehicular networks. 

\section{System Model}

\begin{figure}
\centering
\includegraphics[width=0.7\linewidth]{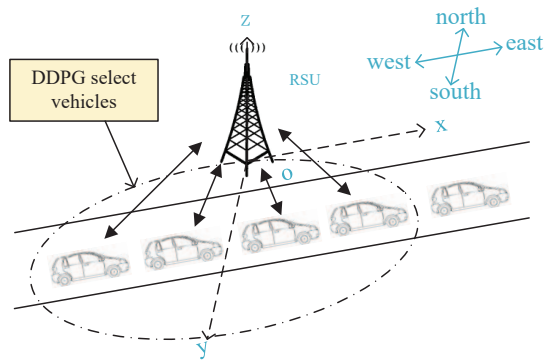}
\caption{System model.} \label{fig1}
\end{figure}

This section will describe the system model.
As shown in Fig. ~\ref{fig1}, we consider an edge-assisted vehicular network consisting of a RSU and $K$ vehicles within its coverage. In the network, the bottom of the RSU stands for the origin, the $x$-axis direction is toward east, the $y$-axis direction is toward south, and the $z$-axis direction is perpendicular to the $x$-axis and $y$-axis and along the direction of the RSU's antenna. Specifically, vehicles are assumed to move toward east with the same velocity in the coverage of the RSU, which can match most of cases in the highway scenarios. The time domain is divided into discrete time slots. Each vehicle $i\left( 1\le i\le K \right)$ carries a different amount of data $D_i$ and has different computing capabilities. At the same time, the vehicle mobility incur a time-varying channel condition.

We first use the DRL algorithm to select the vehicles participating in AFL according to the vehicle's transmission rate, amount of available computing resources, and the location of vehicle, and then the selected vehicles will train and upload their local models to the RSU. 
That is, each selected vehicle can use its local data to train the local model, and then the weight of local model is optimized according to the local training delay and transmission delay.
All selected vehicles can asynchronously upload the local models to the RSU. After multiple rounds of model aggregations, we can get a more accurate global model at the RSU. For ease of understanding, the main notations used in this paper are listed in Table \ref{tab1}.

\begin{table}
\caption{Notations used in this paper}
\label{tab1}
\begin{tabular}{|l|l|}
\hline
Notation & Description\\
\hline
$K$ & Total number of vehicles within the coverage of RSU \\
\hline
$v$ & Velocity of vehicle \\
\hline
$D_i$ & The amount of data carried by vehicle $i$ \\
\hline
${{\mu }_{i}}$ & Computing resources of vehicle $i$ \\
\hline
${{P}_{i}}\left( t \right)$ & Position of vehicle $i$ at time slot $t$ \\
\hline
${{d}_{ix}}\left( t \right)$ & Position of vehicle $i$ along the $x$-axis from the antenna of the RSU at time slot $t$ \\
\hline
${{d}_{y}}$ & Position of vehicle $i$ along the $y$-axis from the antenna of the RSU at time slot $t$ \\
\hline
${{d}_{i0}}$ & Initial position of vehicle $i$ along the $x$-axis \\
\hline
${{H}_{r}}$ & Height of RSU's antenna \\
\hline
${{P}_{r}}$ & Position of RSU's antenna \\
\hline
${{d}_{i}}\left( t \right)$ & Distance from vehicle $i$ to the antenna of RSU at time slot $t$ \\
\hline
$t{{r}_{i}}\left( t \right)$ & Transmission rate of vehicle $i$ at time slot $t$ \\
\hline
$B$ & Transmission bandwidth \\
\hline
${{p}_{0}}$ & Transmission power of each vehicle \\
\hline
${{h}_{i}}\left( t \right)$ & Channel gain \\
\hline
$\alpha$ & Path loss exponent \\
\hline
${{\sigma }^{2}}$ & Power of noise \\
\hline
${{\rho }_{i}}$ & Normalized channel correlation coefficient between consecutive time slots \\
\hline
$f_{d}^{i}$ & Doppler frequency of vehicle $i$ \\
\hline
$\Lambda$ & Wavelength \\
\hline
$\theta$ & Angle between the moving direction and the uplink communication direction \\
\hline
${{C}_{0}}$ & Number of CPU cycles required to train a data \\
\hline
$T_{l}^{i}$ & Delay of local training of vehicle $i$ \\
\hline
$T_{u}^{i}\left( t \right)$ & Transmission delay for vehicle $i$ to upload local model at time slot $t$ \\
\hline
$\left| w \right|$ & Size of the local model of each vehicle \\
\hline
$\gamma$ & Discounted factor \\
\hline
$N$ & Total number of time slots \\
\hline
$\delta$ & Parameter of actor network \\
\hline
${{\delta }^{*}}$ & Optimized parameter of actor network \\
\hline
$\xi$ & Parameter of critic network \\
\hline
${{\xi }^{*}}$ & Optimized parameter of critic network \\
\hline
${{\delta }_{1}}$ & Parameter of target actor network \\
\hline
$\delta _{1}^{*}$ & Optimized parameter of target actor network \\
\hline
${{\xi }_{1}}$ & Parameter of target critic network \\
\hline
$\xi _{1}^{*}$ & Optimized parameter of target critic network \\
\hline
$\tau$ & Update parameter for target networks \\
\hline
${{R}_{b}}$ & Replay buffer \\
\hline
${{\Delta }_{t}}$ & The exploration noise at time slot $t$ \\
\hline
$I$ & Size of mini-batch \\
\hline
${{\mu }_{\delta }}$ & Policy approximated by actor network \\
\hline
${{\mu }^{*}}$ & Optimal policy of system \\
\hline
${E}_{\max }$ & Max episode of training stage \\
\hline
$K_l$ & Number of selected vehicles \\
\hline
$l$ & Number of local training \\
\hline
$m_1$ & Parameter of training weight \\
\hline
$m_2$ & Parameter of transmission weight \\
\hline
$E_{\max }^{'}$ & Max episode of testing stage \\
\hline
\end{tabular}
\end{table}

\section{Parameters Computing}
For simplicity, we first introduce some parameters used in the following sections.

\subsection{Local Training Delay}
Vehicle $i$ uses its local data to train a local model, thus the local training delay $T_{l}^{i}$ of vehicle $i$ can be calculated as:
\begin{equation}
T_{l}^{i}=\frac{{{D}_{i}}{{C}_{0}}}{{{\mu }_{i}}}
\label{eq1}
\end{equation}
where $C_0$ is the number of CPU cycles required to process one unit of data, ${{\mu }_{i}}$ is the computing resources of vehicle $i$, i.e., CPU cycles frequency.

\subsection{Distance}

Denote ${{P}_{i}}\left( t \right)$ as the position of vehicle $i$ at time slot $t$, ${{d}_{ix}}\left( t \right)$ and $d_y$ as the distance between vehicle $i$  and the antenna of RSU at time slot $t$ along the $x$-axis and $y$-axis. Thus ${{P}_{i}}\left( t \right)$ can be expressed as $\left( {{d}_{ix}}\left( t \right),{{d}_{y}},0 \right)$.
Here, $d_y$ is a fixed value, and ${{d}_{ix}}\left( t \right)$ can be denoted as:
\begin{equation}
{{d}_{ix}}\left( t \right)={{d}_{i0}}+vt
\label{eq2}
\end{equation}
where $d_{i0}$ is the initial position of vehicle $i$ along $x$-axis.

We set the height of RSU's antenna as $H_r$, and thus the position of antenna of RSU can be expressed as ${{P}_{r}}=\left( 0,0,{{H}_{r}} \right)$. Then the distance between vehicle $i$ and the antenna of RSU at time slot $t$ can be expressed as:
\begin{equation}
{{d}_{i}}\left( t \right)=\left\| {{P}_{i}}\left( t \right)-{{P}_{r}} \right\|
\label{eq3}
\end{equation}

\subsection{Transmission Rate}
We set the transmission rate of vehicle $i$ at time slot $t$ to be $t{r_i}\left( t \right)$. According to Shannon's theorem, it can be expressed as:
\begin{equation}
t{r_i}\left( t \right) = B{\log _2}\left( {1 + {{{p_0}{\rm{\cdot}}{h_i}\left( t \right){\rm{\cdot}}{{\left( {{d_i}\left( t \right)} \right)}^{ - \alpha }}} \over {{\sigma ^2}}}} \right)
\label{eq4}
\end{equation}
where $B$ is the transmission bandwidth, $p_0$ is the transmission power of each vehicle, ${{h}_{i}}\left( t \right)$ is the channel gain of vehicle $i$ at time slot $t$, $\alpha$ is the path loss exponent, ${{\sigma }^{2}}$ is the power of noise.

We use autoregressive model to formulate the relationship between $h_i\left( t \right)$ and $h_i\left( t-1 \right)$:
\begin{equation}
{{h}_{i}}\left( t \right)={{\rho }_{i}}{{h}_{i}}\left( t-1 \right)+e\left( t \right)\sqrt{1-\rho _{i}^{2}}
\label{eq5}
\end{equation}
where ${{\rho }_{i}}$ is the normalized channel correlation coefficient between consecutive time slots, $e\left( t \right)$ is the error vector following a complex Gaussian distribution and is related to ${{h}_{i}}\left( t \right)$.
According to Jake's fading spectrum, ${{\rho }_{i}}={{J}_{0}}\left( 2\pi f_{d}^{i}t \right)$, where ${J_0}\left( {\rm{\cdot}} \right)$ is the zeroth-order Bessel function of the first kind and $f_{d}^{i}$ is the Doppler frequency of vehicle $i$ which can be calculated as:
\begin{equation}
f_{d}^{i}=\frac{v}{\Lambda }\cos \theta
\label{eq6}
\end{equation}
where $\Lambda$ is the wavelength, $\theta$ is the angle between the moving direction ${x_0} = \left( {1,0,0} \right)$ and the uplink communication direction ${P_r} - {P_i}\left( t \right)$. Thus $\cos \theta$ can be calculated as:
\begin{equation}
\cos \theta  = {{{x_0}{\rm{\cdot}}\left( {{P_r} - {P_i}\left( t \right)} \right)} \over {\left\| {{P_r} - {P_i}\left( t \right)} \right\|}}
\label{eq7}
\end{equation}

\subsection{Transmission Delay}
The transmission delay of vehicle $i$ for uploading its local model $T_{u}^{i}\left( t \right)$ can be denoted as:
\begin{equation}
T_{u}^{i}\left( t \right)=\frac{\left| w \right|}{t{{r}_{i}}\left( t \right)}
\label{eq8}
\end{equation}
where $\left| w \right|$ is the size of the local model of each vehicle.

\section{Problem Formulation}
In this section, we will formulate the problem and define the state, action and reward.

In the system, due to the mobility and the time-varying computing resources and channel conditions of the vehicles, we employ a DRL framework including state, action, and reward to formulate the problem of vehicle selection. Specifically, at each time slot $t$, the system takes action according to the policy based on the current state, and then gets the reward and transitions to the next state. Next, the state, action and reward of the system will be defined, respectively.

\subsection{State}
Considering that the vehicle mobility can be reflected by its position while the local training delay and transmission delay of the vehicle are related to the vehicle's time-varying available computing resources and current channel condition, so we define the state at time slot $t$ as:
\begin{equation}
s\left( t \right)=\left( Tr\left( t \right),\mu \left( t \right),{{d}_{x}}\left( t \right),a\left( t-1 \right) \right)
\label{eq9}
\end{equation}
where $Tr\left( t \right)$ is the set of the transmission rates of each vehicle at time slot $t$, i.e., $Tr\left( t \right)=\left( t{{r}_{1}}\left( t \right),t{{r}_{2}}\left( t \right),\ldots t{{r}_{K}}\left( t \right) \right)$,
$\mu \left( t \right)$ is the set of available computing resources of all vehicles at time slot $t$, i.e., $\mu \left( t \right)=\left( {{\mu }_{1}}\left( t \right),{{\mu }_{2}}\left( t \right),\ldots {{\mu }_{K}}\left( t \right) \right)$,
${{d}_{x}}\left( t \right)$ is the set of all vehicles' positions along the $x$-axis at time slot $t$, i.e., ${{d}_{x}}\left( t \right)=\left( {{d}_{1x}}\left( t \right),{{d}_{2x}}\left( t \right),\ldots {{d}_{Kx}}\left( t \right) \right)$,
$a\left( t-1 \right)$ is the action at time slot $t-1$.

\subsection{Action}
Since the purpose of DRL is to select the better vehicles for AFL according to the current state, we define the system action at time slot $t$ as:
\begin{equation}
a\left( t \right)=\left( {{\lambda }_{1}}\left( t \right),{{\lambda }_{2}}\left( t \right),\ldots {{\lambda }_{K}}\left( t \right) \right)
\label{eq10}
\end{equation}
where ${\lambda _i}\left( t \right),i \in \left[ {1,K} \right]$ is the probability of selecting vehicle $i$, and we define that ${\lambda _1}\left( 0 \right) = {\lambda _2}\left( 0 \right) =  \ldots  = {\lambda _K}\left( 0 \right) = 1$.

We denote a new set ${a_d}\left( t \right) = \left( {{a_{d1}}\left( t \right),{a_{d2}}\left( t \right), \ldots {a_{dK}}\left( t \right)} \right)$ in order to select specific vehicles.
After we normalize the action, if the value of ${{\lambda }_{i}}\left( t \right)$ is greater than or equal to 0.5, ${{a}_{di}}\left( t \right)$ is recorded as 1, otherwise 0. Then we get the set that is composed of of 0 and 1 where the binary value stands for if a vehicle is selected or not.

\subsection{Reward}
We aim to select a vehicle with better performance for AFL to obtain a more accurate global model at the RSU where both the local training delay, transmission delay and the accuracy of the global model are critical metrics. Thus, we define the system reward at time slot $t$ as:
\begin{equation}
r\left( t \right)=-\frac{K}{\sum\limits_{i=1}^{K}{{{\lambda }_{i}}\left( t \right)}}\left[ {{\omega }_{1}}Loss\left( t \right)+{{\omega }_{2}}\frac{\sum\limits_{i=1}^{K}{\left( T_{l}^{i}+T_{u}^{i}\left( t \right) \right){{a}_{di}}\left( t \right)}}{\sum\limits_{i=1}^{K}{{{a}_{di}}\left( t \right)}} \right]
\label{eq11}
\end{equation}
where ${\omega}_{1}$ and ${\omega}_{2}$ are the non-negative weighting factors, $Loss\left( t \right)$ is the loss computed by AFL, which will be discussed later.

The expected long-term discount reward of the system can be expressed as:
\begin{equation}
J\left( \mu  \right)=E\left[ \sum\limits_{t=1}^{N}{{{\gamma }^{t-1}}r\left( t \right)} \right]
\label{eq12}
\end{equation}
where $\gamma \in \left( 0,1 \right)$ is the discounted factor, $N$ is the total number of time slots, $\mu$ is the policy of system.
In this paper we aim to find an optimal policy to maximize the expected long-term discounted reward of the system.

\section{DRL-based AFL weight Optimization:DAFL}
In this section, we introduce the overall system framework, and the training stage to obtain the optimal strategy, then present the testing stage for the performance evaluation of our model.

\subsection{Training Stage}

Considering the state and action spaces are continuous and DDPG is suitable for solving DRL problems under continuous state and action spaces, we employ DDPG to solve our problem.

DDPG algorithm is based on actor-critic network architecture. The actor network is used for policy promotion, and the critic network is used for policy evaluation. Here, both actor and critic network are constructed by deep neural network (DNN). Specifically, actor network is used to approximate policy $\mu$, and the approximated policy is expressed as ${\mu}_{\delta}$.
The actor network observes state and output the action based on policy ${\mu}_{\delta}$.

We improve and evaluate the policy iteratively in order to obtain the optimal policy. To ensure the stability of the algorithm, the target network composed of the target actor network and the target critic network are also employed in the DDPG, where the architectures are the same as the original actor and critic networks, respectively. The proposed algorithm is shown in Algorithm \ref{al1}.

\begin{algorithm}
  \caption{Training Stage for the DAFL-based Framework}
  \label{al1}
  \KwIn{$\gamma$, $\tau$, $\delta$, $\xi$, $a\left(0\right)=\left({1,1,\ldots,1} \right)$}
  \KwOut{optimized ${\delta }^{*}$, $\xi^{*}$}
  Randomly initialize the $\delta$, $\xi$\;
  Initialize target networks by ${{\delta }_{1}}\leftarrow \delta$, ${{\xi }_{1}}\leftarrow \xi$\;
  Initialize replay buffer $R_b$\;
  \For{episode from $1$ to $E_{max}$ }
  {
    Reset simulation parameters of system model, initialize the global model at the RSU\;
    Receive initial observation state $s\left( 1 \right)$\;
    \For{time slot $t$ from $1$ to $N$ }
    {
      Generate the action according to the current policy and exploration noise $a={{\mu }_{\delta }}\left( s|\delta  \right)+{{\Delta }_{t}}$ \;
      Compute $a_d$, get the selected vehicles\;
      The selected vehicles conduct AFL based on weight to train global model at RSU\;
      Get the reward $r$ and the next state $s'$\;
      Store transition $\left( s,a,r,s' \right)$ in $R_b$\;
      \If {number of tuples in $R_b$ is larger than $I$}
      {
      Randomly sample a mini-batch of $I$ transitions tuples from $R_b$\;
      Update the critic network by minimizing the loss function according to Eq. \eqref{eq16}\;
      Update the actor network according to Eq. \eqref{eq17}\;
      Update target networks according to Eqs. \eqref{eq18} and \eqref{eq19}.
      }
    }
}
\end{algorithm}

Let $\delta$ be the actor network parameter, $\xi$ be the critic network parameter, ${{\delta}^{*}}$ be the optimized actor network parameter, ${{\xi}^{*}}$ be the optimized critic network parameter, ${{\delta }_{1}}$ be the target actor network parameter and ${{\xi}_{1}}$ be the target critic network parameter.
$\tau$ is the update parameter of target network and ${{\Delta}_{t}}$ is the exploration noise at time slot $t$. $I$ is the size of mini-batch. Now, we will describe our algorithm in detail.

First, we initialize $\delta$ and $\xi$ randomly, and initialize $\delta_1$ and $\xi_1$ in the target network as $\delta$ and $\xi$ respectively. At the same time, we initialize the replay buffer $R_b$.

Then our algorithm will be executed for $E_{max}$ episodes.
In the first episode, we first initialize the positions of all vehicles, the channel states and computing resources of vehicles, and set ${{\lambda }_{1}}\left( 0 \right)\text{=}{{\lambda }_{2}}\left( 0 \right)\text{=}\ldots \text{=}{{\lambda }_{K}}\left( 0 \right)\text{=}1$. Then at time slot 1, the system can get the state, i.e., $s\left( 1 \right)=\left( Tr\left( 1 \right),\mu \left( 1 \right),{{d}_{x}}\left( 1 \right),a\left( 0 \right) \right)$.
Meanwhile, the convolutional neural network (CNN) is employed as the global model ${{w}_{0}}$ at the RSU.

Our algorithm will execute from time slot 1 to time slot $N$.
At time slot 1, actor network can get the output ${{\mu}_{\delta}}\left( s|\delta  \right)$ according to the state. Note that we add a random noise ${{\Delta}_{t}}$ to the action, and the system can get the action as $a\left( 1 \right)={{\mu}_{\delta }}\left( s\left( 1 \right)|\delta  \right)+{{\Delta}_{t}}$.
Thus we get ${{a}_{d}}\left( 1 \right)$ based on the action and determine the selected vehicles at this time slot. The selected vehicles will conduct AFL. That is, all the selected vehicles train their local models according to their local data, then upload them to RSU asynchronously for global model updating.
Given the action, we can get the reward at time slot 1.
After that we update the positions of vehicles according to Eq. \eqref{eq2}, recalculate the channel states and the available computing resources of the vehicles, and update the transmission rates of the vehicles according to Eq. \eqref{eq4}. Then it can get the next state $s\left( 2 \right)$.
The related sample $\left( s\left( 1 \right),a\left( 1 \right),r\left( 1 \right),s\left( 2 \right) \right)$ will be stored in $R_b$.
Note that the system will iteratively calculate and store the samples into $R_b$, until reaching the capacity of $R_b$.

When the number of tuples is bigger than $I$ in $R_b$, the parameters $\delta$, $\xi$, $\delta_1$ and $\xi_1$ of actor network, critic network and target networks respectively will be trained to maximize $J\left( {{\mu}_{\delta}} \right)$.
Here, $\delta$ is updated towards the gradient direction of $J\left( {{\mu}_{\delta}} \right)$, i.e., ${{\nabla }_{\delta }}J\left( {{\mu }_{\delta }} \right)$.
We set ${{Q}_{{{\mu}_{\delta}}}}\left( s\left( t \right),a\left( t \right) \right)$ as the action-value function which obeys policy ${{\mu }_{\delta }}$ under $s\left( t \right)$ and $a\left( t \right)$, it can be expressed as:
\begin{equation}
{{Q}_{{{\mu }_{\delta }}}}\left( s\left( t \right),a\left( t \right) \right)={{E}_{{{\mu }_{\delta }}}}\left[ \sum\limits_{{{k}_{1}}=t}^{N}{{{\gamma}^{{{k}_{1}}-t}}r\left( {{k}_{1}} \right)} \right]
\label{eq13}
\end{equation}
It represents the long term expected discount reward at time slot $t$.

The existing paper has proved that the solution to ${{\nabla }_{\delta }}J\left( {{\mu }_{\delta }} \right)$ can be replaced by solving the gradient of ${{Q}_{{{\mu }_{\delta }}}}\left( s\left( t \right),a\left( t \right) \right)$, i.e., ${{\nabla }_{\delta }}{{Q}_{{{\mu }_{\delta }}}}\left( s\left( t \right),a\left( t \right) \right)$ \cite{s1}.
Due to the continuous action space of ${{Q}_{{{\mu }_{\delta }}}}\left( s\left( t \right),a\left( t \right) \right)$, it cannot be solved by Bellman Equation \cite{s2}.
In order to solve this problem, critic network uses $\xi$ to approximate ${{Q}_{{{\mu }_{\delta }}}}\left( s\left( t \right),a\left( t \right) \right)$ by ${{Q}_{\xi }}\left( s\left( t \right),a\left( t \right) \right)$.

When the number of tuples is bigger than $I$ in $R_b$, system will sample $I$ tuples randomly from $R_b$ to form a mini-batch. Let $\left( {{s}_{x}},{{a}_{x}},{{r}_{x}},s_{x}^{'} \right),x\in \left[ 1,2,\ldots ,I \right]$ be the $x$-th tuple in the mini-batch.
The system will input the $s_{x}^{'}$ to the target actor network, and get the output action $a_{x}^{'}={{\mu }_{{{\delta }_{1}}}}\left( s_{x}^{'}|{{\delta }_{1}} \right)$.
Then input $s_{x}^{'}$ and $a_{x}^{'}$ to the target critic network, and get the action-value function ${{Q}_{{{\xi }_{1}}}}\left( s_{x}^{'},a_{x}^{'} \right)$.
The target value can be calculated as:
\begin{equation}
{{y}_{x}}={{r}_{x}}+\gamma {{Q}_{{{\xi }_{1}}}}\left( s_{x}^{'},a_{x}^{'} \right){{|}_{a_{x}^{'}={{\mu }_{{{\delta }_{1}}}}\left( s_{x}^{'}|{{\delta }_{1}} \right)}}
\label{eq14}
\end{equation}
According to ${{s}_{x}}$ and ${{a}_{x}}$, critic network will output the ${{Q}_{\xi }}\left( {{s}_{x}},{{a}_{x}} \right)$, then the loss of tuple $x$ is given by:
\begin{equation}
{{L}_{x}}={{\left[ {{y}_{x}}-{{Q}_{\xi }}\left( {{s}_{x}},{{a}_{x}} \right) \right]}^{2}}
\label{eq15}
\end{equation}
When all the tuples act as the input to the critic network and target networks, we can get the loss function:
\begin{equation}
L\left( \xi  \right)=\frac{1}{I}\sum\limits_{x=1}^{I}{{{L}_{x}}}
\label{eq16}
\end{equation}
In this case, the critic network updates $\xi$ by employing the gradient descent of ${{\nabla }_{\xi }}L\left( \xi  \right)$ to the loss function $L\left( \xi  \right)$.

Similarly, actor network updates $\delta$ by employing the gradient ascent, i.e., ${{\nabla}_{\delta}}J\left( {{\mu}_{\delta}} \right)$, to minimize $J\left( {{\mu}_{\delta}} \right)$\cite{s3}, where ${{\nabla }_{\delta }}J\left( {{\mu }_{\delta }} \right)$ is calculated by action-value function approximated by critic network as follows:
\begin{equation}
\begin{split}
&\nabla_{\delta}J(\mu_{\delta})\\
&\approx \frac{1}{I}\sum_{x=1}^{I}\nabla_{\delta}Q_{\xi}(s_x,a_{x}^{\mu})|_{a_{x}^{\mu}=\mu_{\delta}(s_x|\delta)}\\
&=\frac{1}{I}\sum_{x=1}^{I}\nabla_{a_{x}^{\mu}}Q_{\xi}(s_x,a_{x}^{\mu})|_{a_{x}^{\mu}=\mu_{\delta}(s_x|\delta)} \\&\quad\cdot \nabla_{\delta}\mu_{\delta}(s_x|\delta)
\end{split}
\label{eq17}
\end{equation}
Here the input of ${Q}_{\xi}$ is $a_{x}^{\mu}={{\mu}_{\delta}}\left({{s}_{x}}|\delta  \right)$.

In the end of a time slot $t$, we update the parameters of target networks as follows:
\begin{equation}
{{\xi}_{1}}\leftarrow \tau \xi +\left( 1-\tau  \right){{\xi}_{1}}
\label{eq18}
\end{equation}
\begin{equation}
{{\delta}_{1}}\leftarrow \tau \delta +\left( 1-\tau  \right){{\delta}_{1}}
\label{eq19}
\end{equation}
where $\tau$ is a constant and $\tau \ll 1$.

\begin{figure}
\center
\includegraphics[width=\textwidth]{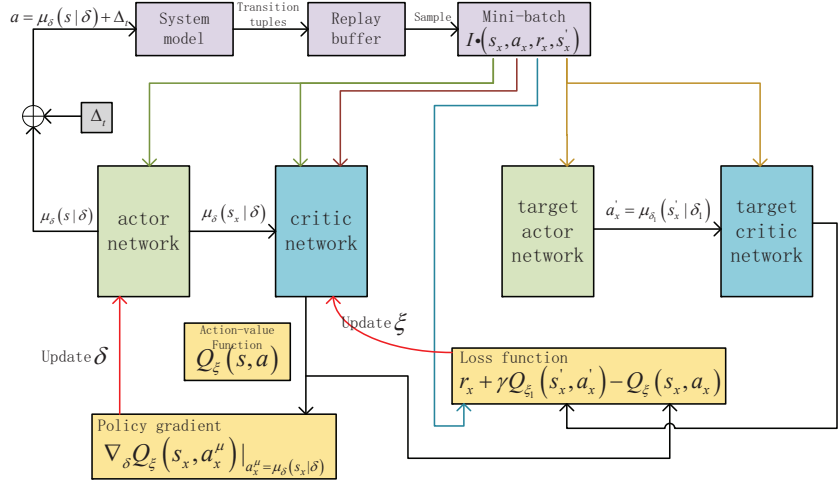}
\caption{DDPG flow diagram}
\label{fig2}
\end{figure}

Then input $s'$ to actor network and start the same procedure for the next time slot.
When the time slot $t$ reaches $N$, this episode is completed.
In this case, system will initialize the state $s\left( 1 \right)=\left( Tr\left( 1 \right),\mu \left( 1 \right),{{d}_{x}}\left( 1 \right),a\left( 0 \right) \right)$ and execute the next episode.
When the number of episodes reaches $E_{max}$, the training is finished. 
We get the optimized ${\delta }^{*}$, ${\xi }^{*}$, $\delta _{1}^{*}$ and $\xi _{1}^{*}$.
The overall DDPG flow diagram is listed in Fig.~\ref{fig2}.

\subsection{Process of AFL}
In this section, we will introduce the process of AFL in detail, which is used in the step 10 in Algorithm \ref{al1}.

Let ${{V}_{k}},k\in \left[ 1,{{K}_{1}} \right]$ be the selected vehicles, where $K_l$ is the total number of the selected vehicles.
In the AFL, each vehicle will go through three stages: global model downloading, local training, uploading and updating.
Specifically, vehicle $V_k$ will first download the global model from the RSU, then it will train a local model using local data for some iterations. Then it will upload the local model to the RSU. Once the RSU receives a local model, it updates the global model immediately.
To be clearly, we use the AFL training at time slot $t$ of vehicle $V_k$ as an example.

\subsubsection{Downloading the Global Model}
In time slot $t$, vehicle $V_k$ downloads the global model ${w}_{t-1}$ from the RSU. Note that  the global model at the RSU is initialized as $w_0$ using CNN at the beginning of the whole training process.

\subsubsection{Local Training}
Vehicle $V_k$ trains local model (CNN) based on its local data. The local training includes $l$ iterations.
In iteration $m\left( m\in \left[ 1,l \right] \right)$, vehicle $V_k$ first inputs the data $a$ into the CNN of local model ${{w}_{k,m}}$, then outputs the prediction probability $\hat{{y}_{a}}$ of each label $y_a$ of data $a$.
Cross-entropy loss function is used to compute the loss of ${{w}_{k,m}}$:
\begin{equation}
{{f}_{k}}\left( {{w}_{k,m}} \right)\text{=}-\sum\limits_{a=1}^{{{D}_{i}}}{{{y}_{a}}\log }\hat{y_a}\,
\label{eq20}
\end{equation}
Then stochastic gradient descent (SGD) algorithm is used to update our model as follows:
\begin{equation}
{{w}_{k,m+1}}={{w}_{k,m}}-\eta \nabla {{f}_{k}}\left( {{w}_{k,m}} \right)
\label{eq21}
\end{equation}
where $\nabla {{f}_{k}}\left( {{w}_{k,m}} \right)$ is the gradient of ${{f}_{k}}\left( {{w}_{k,m}} \right)$, $\eta$ is the learning rate. Vehicle $V_k$ will use the updated local model in the proceeding iteration of $m+1$.
The local training will stop when the iteration reaches $l$. At this time, the vehicle gets the updated local model $w_k$.

For local model $w_k$, the loss is:
\begin{equation}
{{f}_{k}}\left( {{w}_{k}} \right)\text{=}-\sum\limits_{a=1}^{{{D}_{i}}}{{{y}_{a}}\log }\hat{y_a}\,
\label{eq22}
\end{equation}

In our proposed scheme, the impact of the delay on the model has also been investigated.  Specifically, the local training and local model uploading will incur some delay, during which  other vehicles may upload the local models to the RSU.
In this situation, the local model of this vehicle will have staleness.
Considering this issue, we introduce the training weight and transmission weight.

The training weight is related to the local training delay, and it can be expressed as:
\begin{equation}
{{\beta }_{1,k}}={{m}_{1}}^{T_{l}^{{{V}_{k}}}-0.5}
\label{eq23}
\end{equation}
where $T_{l}^{{{V}_{k}}}$ is the local training delay of vehicle $V_k$, which can be calculated by Eq. \eqref{eq1}. ${{m}_{1}}\in \left( 0,1 \right)$ is the parameter to make ${{\beta }_{1,k}}$ decrease with the increase of local training delay.

The transmission weight is related to the transmission delay of vehicles for uploading local models to the RSU.
Here, due to the downloading delay, i.e., the duration of vehicles downloading the global model from the RSU, is very small compared with transmission delay so it can be ignored, thus the transmission weight can be denoted as:
\begin{equation}
{{\beta }_{2,k}}\left( t \right)={{m}_{2}}^{T_{u}^{{{V}_{k}}}\left( t \right)-0.5}
\label{eq24}
\end{equation}
where $T_{u}^{{{V}_{k}}}\left( t \right)$ is the transmission delay of $V_k$, which can be calculated by Eq. \eqref{eq8}, ${{m}_{2}}\in \left( 0,1 \right)$ is the parameter to make ${{\beta }_{2,k}}$ decrease with the increase of transmission delay.

Then we can get the weight optimized local model, i.e., 
\begin{equation}
{{w}_{kw}}={{w}_{k}}*{{\beta }_{1,k}}*{{\beta }_{2,k}}
\label{eq25}
\end{equation}

\subsubsection{Uploading and Updating}
When vehicle $V_k$ uploads the weight optimized local model, the RSU will update the global model. The formula is
\begin{equation}
{{w}_{new}}=\beta {{w}_{old}}+\left( 1-\beta  \right){{w}_{kw}}
\label{eq26}
\end{equation}
where ${{w}_{old}}$ is the current global model at the RSU, $w_{new}$ is the updated global model, $\beta \in \left( 0,1 \right)$ is the proportion of aggregation.

When RSU receives the first uploaded weight optimized local model, ${{w}_{old}}={{w}_{t-1}}$.
When RSU receives all the weight optimized local models of selected vehicles and get the global model $w_t$ updated for $K_l$ rounds, it indicates the global model updating is finished at this time slot.

At the same time, we get the average loss of the selected vehicles:
\begin{equation}
Loss\left( t \right)=\frac{1}{{{K}_{l}}}\sum\limits_{k=1}^{{{K}_{l}}}{{{f}_{k}}\left( {{w}_{k}} \right)}
\label{eq27}
\end{equation}
Then step 10 in Algorithm \ref{al1} is completely explained.
The procedure of AFL training is shown in Algorithm \ref{al2}.

\begin{algorithm}
	\small
	\caption{weight optimized AFL scheme}
	\label{al2}
	Initialize the global model $w_0$;\\
	\For{each round $x$ from $1$ to $K_l$}
	{
		$w_k \leftarrow \textbf{Vehicle Updates}(w_0)$;\\
		Vehicle $V_k$ calculates the weight optimized local model $w_{kw}$ based on Eq. \eqref{eq25};\\
		Vehicle $V_k$ uploads the weight optimized local model $w_{kw}$ to the RSU;\\

		RSU receives the weight optimized local model $w_{kw}$;\\
		RSU updates the global model based on Eq. \eqref{eq26};\\
		\Return $w_{new}$
	}
    Get the updated global model $w_t$ after $K_l$ rounds.\\
	\textbf{Vehicle Update}($w$):\\
	\textbf{Input:} $w_0$ \\
	\For{each local iteration $m$ from $1$ to $l$}
	{
		Vehicle $V_k$ calculates the cross-entropy loss function based on Eq. \eqref{eq20};\\
		Vehicle $V_k$ updates the local model based on Eq. \eqref{eq21};\\
	}
	Set $w_k=w_{k,l}$;\\
	\Return$w_k$
	
\end{algorithm}

\subsection{Testing Stage}
The testing stage employs the achieved critic network, target actor network and target critic network in the training stage. In the testing stage, the system will select the policy with optimized parameter ${{\delta }^{*}}$.
The process of testing stage is shown in Algorithm \ref{al3}.

\begin{algorithm}
  \caption{Testing Stage for the DAFL-based Framework}
  \label{al3}
  \For{episode from $1$ to $E_{\max }^{'}$ }
  {
    Reset simulation parameters of system model, initialize the global model at the RSU\;
    Receive initial observation state $s\left( 1 \right)$\;
    \For{time slot $t$ from $1$ to $N$ }
    {
      Generate the action according to the current policy $a={{\mu }_{\delta }}\left( s|\delta  \right)$ \;
      Compute $a_d$, get the selected vehicles\;
      The selected vehicles conduct AFL based on weight to train global model at the RSU\;
      Get the reward $r$ and the next state ${{s}^{'}}$\;
    }
  }
\end{algorithm}

\section{Simulation and Results}

\subsection{Simulation Setup}
In this section, the simulation tool is python 3.9. Our actor network and critic network are all selected as the DNN with two hidden layers. The two hidden layers have 400 and 300 neurons, respectively.
The exploration noise obeys the Ornstein-Uhlenbeck (OU) process with variation 0.02 and decay-rate 0.15.
We use MNIST dataset to allocate data to vehicles and the computing resources of vehicles obey a truncated Gaussian distribution. Here the unit of the computing resources is CPU-cycles/s.
We configured a vehicle as the bad vehicle, that is, it has small amount of data and computing resources, and the local model is disturbed by random noise.
The rest simulation parameters are shown in Table \ref{tab2}.

\begin{table}\footnotesize
\caption{Parameters of simulation}
\label{tab2}
\centering
\begin{tabular}{|l|l|l|l|}
\hline
Parameter &Value &Parameter &Value\\
\hline
$\gamma$ & 0.99 & $H_r$ & 10$m$ \\
\hline
$\tau$ & 0.001 & $B$ & 1000$HZ$ \\
\hline
$I$ & 64 & $p_0$ & 0.25$w$ \\
\hline
$E_{max}$ & 1000 & ${{\sigma }^{2}}$ & ${{10}^{-9}}mw$ \\
\hline
$E_{\max }^{'}$ & 3 & $\Lambda$ & 7$m$ \\
\hline
$K$ & 5 & $\left| w \right|$ & 5000$bits$ \\
\hline
$v$ & 20$m/s$ & $\alpha$ & 2 \\
\hline
$C_0$ & ${{10}^{6}}CPU-cycles$ & $m_1$ & 0.9 \\
\hline
$t$ & 0.5 & $m_2$ & 0.9 \\
\hline
$d_y$ & 5$m$ & $$ &  \\
\hline
\end{tabular}
\end{table}

\subsection{Experiment Results}
\begin{figure}
\centering
\includegraphics[width=0.8\linewidth]{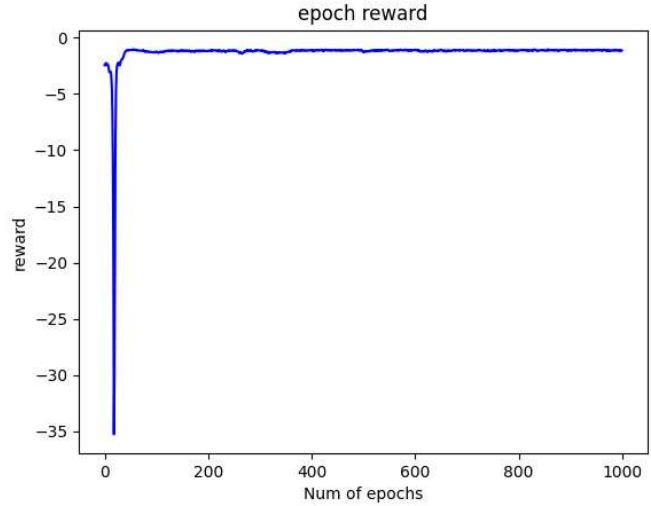}
\caption{Reward for different epochs}
\label{fig3}
\end{figure}
Fig.~\ref{fig3} shows the system reward with respect to different epochs in the training stage.
One can see that when the number of epoches is small, the reward has a large variation. This is due to that the system is learning and optimizing the network in the initial phase, so some explorations (i.e., action) incur poor performance. As the number of epoches increases, the reward gradually becomes stable and smoother. It means that at this time, the system has gradually learned the optimal policy, and the training of the neural network has been close to be completed.

\begin{figure}
\center
\includegraphics[width=0.9\linewidth]{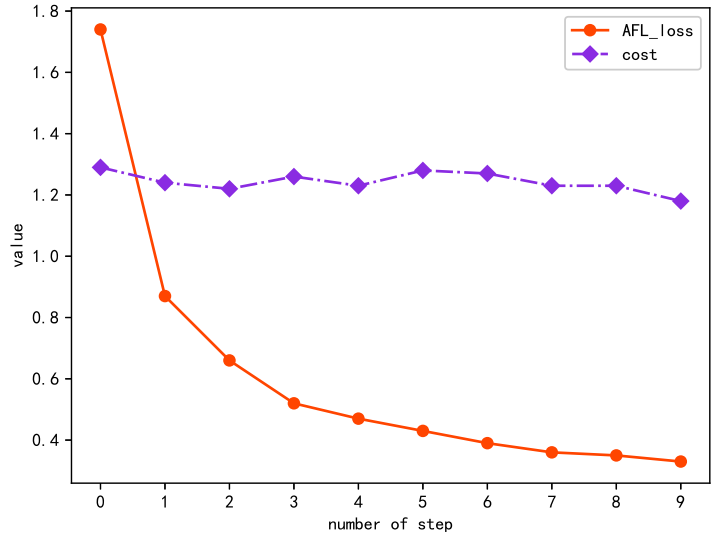}
\caption{Relation between delay and loss}
\label{fig4}
\end{figure}

Fig.~\ref{fig4} depicts the two components of the reward in the testing stage: the loss calculated in the AFL and the sum of the local training delay and the transmission delay.
It can be seen that the loss is decreasing as the number of steps increases.
This is attributed to the fact that vehicles are constantly uploading local model to update global model at the RSU, so the global model becomes more accurate.
The sum of delay shows a certain fluctuation,
because of the dynamic available computing resources of the vehicle and its time-varying location.

\begin{figure}
\center
\includegraphics[width=0.9\linewidth]{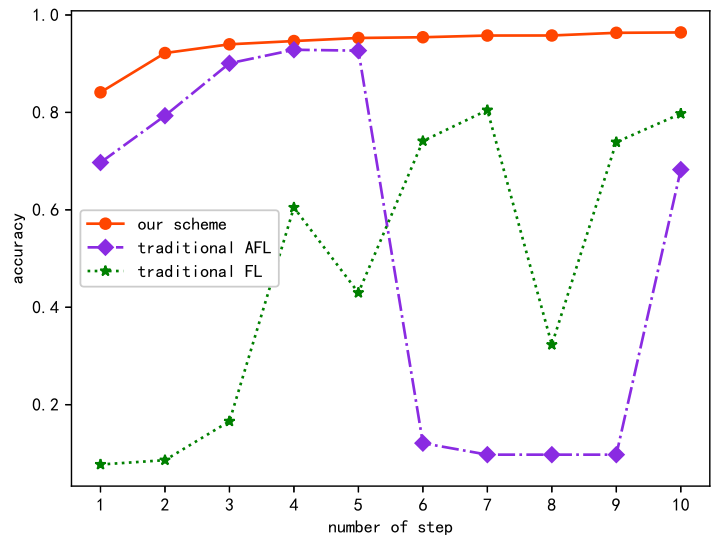}
\caption{Accuracy v.s. number of steps in testing stage}
\label{fig5}
\end{figure}

Fig.~\ref{fig5} shows the accuracy of our scheme, traditional AFL and traditional FL in the presence of bad node.
From the figure one can see that the accuracy of our scheme remains at a good level and gradually increases and finally reaches stability. This indicates that our scheme can effectively remove the bad node in the model training.
Since traditional AFL and FL do not have the function of selecting the vehicles, their accuracy are seriously affected by the bad node, resulting in large fluctuations.

\begin{figure}
\center
\includegraphics[width=0.9\linewidth]{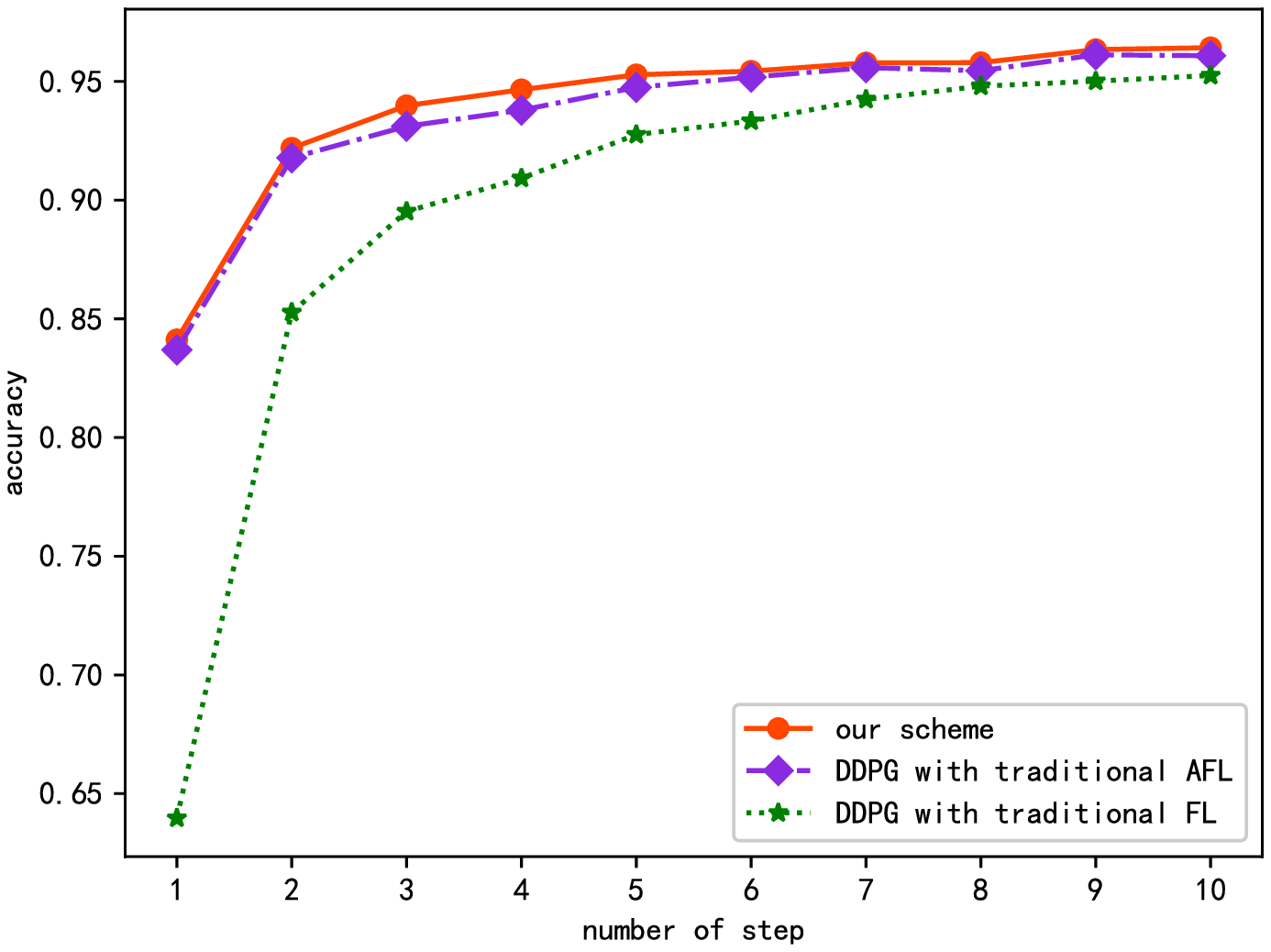}
\caption{Accuracy with optimized model weights in testing stage}
\label{fig7}
\end{figure}

\begin{figure}
\center
\includegraphics[width=0.9\linewidth]{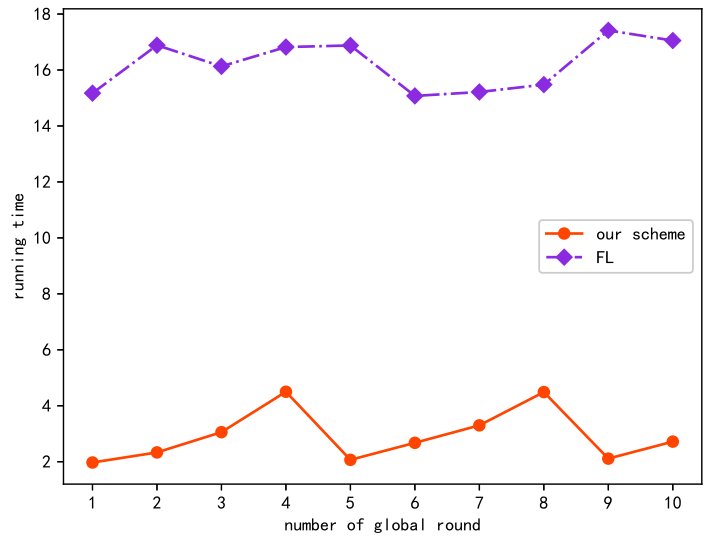}
\caption{Training delay v.s. number of training rounds}
\label{fig9}
\end{figure}

Fig.~\ref{fig7} shows the accuracy of our scheme and traditional AFL and FL after the vehicle selection.
From the figure one can see that the accuracy of all schemes are increasing when the number of step increases and finally become stable. However, our scheme has the highest accuracy among them. This is because our scheme considers the impact of local training delay and transmission delay of vehicles in global model updating.

Fig.~\ref{fig9} depicts the training delay of our scheme compared to the FL as the global round (i.e., step) increases.
It shows the delay of FL remains to be high while our scheme keeps a small delay.
This is because FL only starts updating the global model when all local models of selected vehicles are received, whereas in our scheme, RSU is updated every time when it receives a local model uploaded from a vehicle.
At the same time, we can observe that the delay of our scheme appears to rise at first and then fall and rise again. This is because the proposed scheme selects four vehicles to update the global model one by one.
At the same time, owning to the large local computing delay of the vehicles compared to the transmission delay, the local computing delay of the vehicle dominates. In this case, since the vehicle that finishes the earliest local training will update the global model first, the training delay gradually increases. When the four vehicles all update the global model, the vehicles will repeat the above global model updating until reaches the maximum number of step.

\begin{figure}
\center
\includegraphics[width=0.9\linewidth]{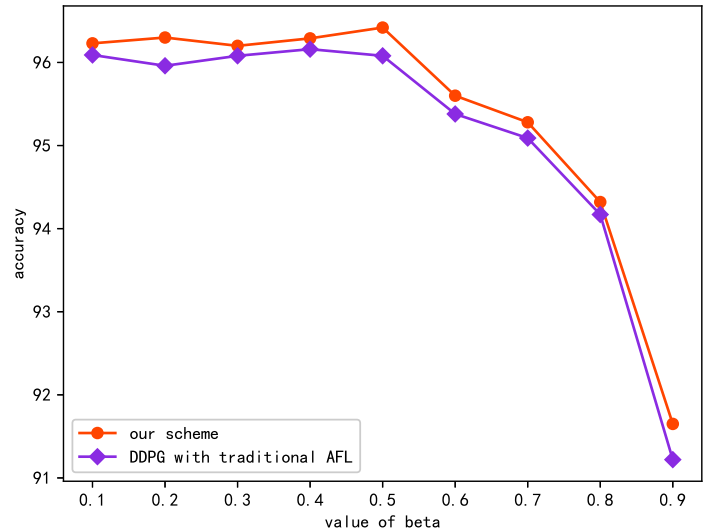}
\caption{Accuracy v.s. different beta}
\label{fig10}
\end{figure}

Fig.~\ref{fig10} depicts the accuracy of our scheme and traditional AFL with selected vehicles under different value of $\beta$.
It shows when $\beta$ is small, the accuracy of the model keeps relatively high. In contrast,  when $\beta$ gradually increases, the accuracy of global model gradually decreases. This is because when $\beta$ is relatively large, the weight of the local model is much small, thus the update of the global model mainly depends on the previous hyperparameter values of global model, which decreases the influence and contribution of new local model of all vehicles, so it has a significant impact on the accuracy of global model.
At the same time, the accuracy of our scheme is better than that of AFL. This is because our scheme has considered the influence of local computing delay and transmission delay of vehicles.

\section{Conclusion}
In this paper, we considered the vehicle mobility, time-varying channel states, time-varying computing resources of vehicles, different amount of local data of vehicles and the situation of bad node and proposed a DAFL-based framework. The conclusions are summarized as follows:
\begin{itemize}
    \item The accuracy of our scheme is better than traditional AFL and FL. This is due to that our scheme can effectively remove the bad node and thus prevent the global model updating from being affected by bad node.

    \item For the absent of bad nodes, the accuracy of our scheme is still higher than that of AFL and FL, because our scheme considers the mobility of the vehicles, time-varying channel conditions, available computing resources of the vehicles and different amounts of local data to allocate different weights to different vehicles's local model in AFL.

	\item The aggregate proportion $\beta$ affects the accuracy of the global model. Specifically,   when $\beta$ is relatively small, it can get the desirable accuracy.
\end{itemize}

%
%
%
%

\end{document}